\documentclass{svmult}
 
\usepackage[T1]{fontenc}
\usepackage[utf8]{inputenc}
\usepackage{amsmath,amssymb,amsfonts,bm}
\usepackage{newtxtext,newtxmath}
\usepackage{makeidx}
\usepackage{multicol}
\usepackage{footmisc}
\usepackage{textcomp}
\usepackage{url}
\usepackage{multirow}
\usepackage{color,soul}
\usepackage{booktabs}
\usepackage{graphicx}
\usepackage{subcaption}

\title*{Towards Fast GNN Surrogates for CO$_2$ Migration in Complex Geological Formations}

\titlerunning{Fast GNN Surrogates for CO$_2$ Migration} 

\author{
Rodrigo S. Luna \and
Thiago H. N. Coelho \and
Luiz S. L. Neto \and
Roberto M. Velho \and
Adriano M. A. Cortes \and
Renato N. Elias \and
Alexandre G. Evsukoff \and
Fernando A. Rochinha \and
Mauricio Araya-Polo \and
Herve Gross \and
Alvaro L. G. A. Coutinho
}
 
\authorrunning{Luna et al.}
 
\institute{
Rodrigo S.~Luna \at
    Systems and Computer Engineering and High Performance Computing Center, NACAD - COPPE, Federal University of Rio de Janeiro; \email{luna@cos.ufrj.br}
\and
Thiago H.~N.~Coelho \at
    Systems and Computer Engineering and High Performance Computing Center, NACAD - COPPE, Federal University of Rio de Janeiro; \email{tcoelho@cos.ufrj.br}
\and
Luiz~S.~L.~Neto \at
    Civil Engineering and High Performance Computing Center, NACAD - COPPE, Federal University of Rio de Janeiro; \email{luizlealn@nacad.ufrj.br}
\and
Roberto M.~Velho \at
    Systems and Computer Engineering and High Performance Computing Center, NACAD - COPPE, Federal University of Rio de Janeiro; \email{roberto.velho@nacad.ufrj.br}
\and
Adriano M.~A.~Cortes \at
    Systems and Computer Engineering and High Performance Computing Center, NACAD - COPPE, Federal University of Rio de Janeiro; \email{adriano@nacad.ufrj.br}
\and
Renato N.~Elias \at
    Civil Engineering and High Performance Computing Center, NACAD - COPPE, Federal University of Rio de Janeiro;  \email{rnelias@nacad.ufrj.br}
\and
Alexandre G.~Evsukoff \at
    Civil Engineering and High Performance Computing Center, NACAD - COPPE, Federal University of Rio de Janeiro;  \email{alexandre.evsukoff@coc.ufrj.br}
\and
Fernando A.~Rochinha \at
    Mechanical Engineering and High Performance Computing Center, NACAD - COPPE, Federal University of Rio de Janeiro; \email{faro@mecanica.coppe.ufrj.br}
\and
Mauricio Araya-Polo \at
    TotalEnergies \email{mauricio.araya@totalenergies.com}
\and
Herve Gross \at
    TotalEnergies \email{herve.gross@totalenergies.com}
\and
Alvaro L.~G.~A.~Coutinho \at
    Civil Engineering and High Performance Computing Center, NACAD - COPPE, Federal University of Rio de Janeiro; \email{alvaro@nacad.ufrj.br}
}
\begin{document}
\maketitle
\abstract{This chapter discusses how a data-driven machine learning approach can reproduce key aspects of the physical behavior of multiphase flows in complex geological formations. We propose an end-to-end graph neural surrogate tailored to CO$_2$ plume migration forecasting in geological storage. The method is evaluated on the SPE11A benchmark, a well-known industry test case designed to assess CO$_2$ storage scenarios and characterized by sharp gas-water interfaces, strong advective transport, and rapid convective mixing with fingering development. The benchmark is reformulated as a graph in which nodes represent computational cells and edges encode transmissibility-based interactions enriched with geometric attributes. Directional transport arising from grid geometry, permeability contrasts, and geological heterogeneity is captured through an anisotropic message-passing mechanism, where interaction weights are computed via geometry-conditioned edge embeddings, biasing message aggregation toward physically relevant transport directions. Temporal evolution is modeled in latent space using an autoregressive residual formulation trained with multi-step supervision. The proposed model produces competitive forecasts of gas saturation and liquid-phase density, which are key indicators for CO$_2$ storage monitoring, with cumulative errors that remain moderate over extended forecasting horizons.}

\section{Introduction}
Accurate and efficient forecasting of CO$_2$ plume migration in complex geological formations is critical for the safe deployment of carbon capture and storage (CCS) technologies. While traditional numerical reservoir simulators are physically reliable, they incur prohibitive computational costs for long-term simulations and uncertainty quantification. Recent scientific machine learning approaches, including Fourier Neural Operators, DeepONets, and Graph Neural Networks (GNNs)~\cite{ju2024learning}, have shown strong accuracy and significant speedups in many-query settings such as parameter exploration and scenario analysis, and are increasingly adopted in digital twins. GNNs are particularly well suited for subsurface applications as they naturally operate on unstructured meshes typical of geological reservoirs. However, most graph-based simulators, including MeshGraphNet~\cite{Pfaff2021MeshGraphNet}, rely on isotropic message passing, propagating information uniformly across neighboring cells, which can be limiting for reservoir flow problems dominated by strongly directional transport induced by grid geometry, permeability contrasts, gravity effects, and geological heterogeneity. In this work, we develop an advanced end-to-end GNN pipeline for CO$_2$ plume migration forecasting that introduces an anisotropic message-passing mechanism explicitly modulated by geometric and physical edge attributes, together with a latent temporal dynamics formulation and an autoregressive training strategy to ensure stable long-horizon predictions. The proposed approach is evaluated on the SPE11A benchmark, a two-dimensional laboratory-scale test case inspired by a controlled CO$_2$ storage experiment~\cite{LandaMarban2025, RASMUSSEN2021159, nordbotten202411th}, governed by a fully nonlinear coupled system of partial differential equations combining Darcy flow, component-wise mass conservation, and phase equilibrium constraints. The benchmark poses significant challenges due to sharp gas--water interfaces and the rapid onset of convective mixing with extensive fingering, making accurate forecasting of gas saturation and phase composition particularly demanding. The remainder of this work is organized as follows. Section~\ref{sec:gnn} briefly describes the GNN engine, which builds on the one introduced in \cite{ju2024learning}. In Section \ref{sec:exp}, we describe the experimental setup, including data generation for SPE11A that accounts for random geologic variations, data preparation, the model and training setup, the evaluation protocol, and the forecasting results. The chapter ends with a summary of our main conclusions.

\section{Graph-Based Forecast Model}\label{sec:gnn}

Our forecasting model builds upon MeshGraphNet~\cite{Pfaff2021MeshGraphNet}, an encoder--processor--decoder graph-based neural simulator for mesh-based physical systems. In its original formulation, MeshGraphNet employs isotropic message passing and is trained with one-step supervision before being deployed autoregressively, which can lead to error accumulation in long-horizon reservoir flow simulations~\cite{ju2024learning}.

Reservoir flow exhibits strongly directional transport due to grid geometry, permeability contrasts, and geological heterogeneity. To better capture these effects, we introduce anisotropic message passing inspired by geometric anisotropic formulations~\cite{thurlemann2023anisotropic}. Instead of tensorial multipole representations, we adopt a lightweight directional edge-weighting mechanism in which interaction weights are computed from geometric features, such as relative cell positions and edge attributes, allowing the model to prioritize physically relevant transport directions.

In addition, to capture temporal dependencies beyond one-step transitions and improve stability in long-horizon forecasts, we incorporate a recurrent latent dynamics module based on a GraphConv--LSTM together with a residual state update formulation. These extensions define the proposed \emph{AnisoMeshGraph-LSTM} architecture.

\subsection{Graph Representation of the SPE11A Benchmark}

We recast the SPE11A benchmark as a graph $G=(V,E)$, where each computational cell is represented as a node and edges connect cells with nonzero transmissibility. The transmissibility at each connection between cells $i$ and $j$ is computed as a function of grid geometry and rock permeability and is equal to zero when two cells are separated by a fault \cite{ju2024learning}. For a given set of mesh and permeability, the graph structure remains fixed for all time steps. The properties associated with node $i$ at time $t$ are called a
node feature. The properties corresponding to edge ($i$,$j$)
are referred to as an edge feature, and are independent
of time. 

Node features encode the physical state of the system at each cell, including time-dependent variables and exogenous inputs, while edge attributes capture geometric and physical interaction properties. Although transmissibility is used to define the graph topology, it is not included directly as an edge feature; instead, the displacement vector and Euclidean distance between cell centers serve as geometric proxies that implicitly encode the connectivity structure.

\subsection{Encoder}

Let $X^{t} \in \mathbb{R}^{n \times F_v}$ denote the physical state of the system at time $t$, defined at all graph nodes. Each row $x_i^{t}$ contains time-dependent state variables (e.g., saturation or density), as well as time-invariant parameters and control inputs associated with node $i$.
The simulator provides a trajectory $\{X^{t}\}_{t=1}^{N}$, which serves both as input to the model and as ground truth during supervised training.

Edge attributes are collected in the matrix
$E_{\mathrm{attr}} \in \mathbb{R}^{m \times F_e}$, where each row $e_{ij}$ encodes geometric and physical properties of the interface between nodes $i$ and $j$.
Node and edge features are embedded into latent spaces using multilayer perceptrons (MLPs):
\begin{equation}
z_i^{t,0} = \phi_v^0(x_i^t) \in \mathbb{R}^{d_v}, 
\qquad
z_{ij}^{t,0} = \phi_e^0(e_{ij}) \in \mathbb{R}^{d_e},
\end{equation}
where $d_v$ and $d_e$ denote the node and edge latent dimensions.
Both $\phi_v^0$ and $\phi_e^0$ are MLPs with residual connections.

\subsection{Processor: Anisotropic Message Passing}

The processor applies $L$ layers of message passing.
At each layer $l = 1, \dots, L$, edge and node embeddings are updated sequentially.
To capture directional interactions induced by reservoir connectivity and geological heterogeneity, we employ anisotropic message passing with learnable, edge-dependent interaction weights.

\paragraph{\bf{Edge Update}}

\begin{equation}
z_{ij}^{t,l} =
\phi_e^l\!\left(
\left[
z_{ij}^{t,l-1},\,
z_i^{t,l-1},\,
z_j^{t,l-1}
\right]
\right),
\end{equation}
where $\phi_e^l$ is an MLP with residual connections and ReLU activation. 

\paragraph{\bf{Anisotropic Weights}}

For each directed edge $(i,j)$, a scalar compatibility score is computed as
\begin{equation}\label{eq:anisotropic}
s_{ij}^{t,l} =
\phi_\alpha^l\!\left(
\left[
z_{ij}^{t,l},\,
z_i^{t,l},\,
z_j^{t,l}
\right]
\right),
\end{equation}
and normalized over the neighborhood of node $i$:
\begin{equation}
\alpha_{ij}^{t,l} =
\frac{\exp(s_{ij}^{t,l})}
{\sum_{k \in \mathcal{N}(i)} \exp(s_{ik}^{t,l})}.
\end{equation}

\paragraph{\bf{Node Update}}

\begin{equation}
z_i^{t,l} =
\phi_v^l\!\left(
\left[
z_i^{t,l-1},\,
\sum_{j \in \mathcal{N}(i)} \alpha_{ij}^{t,l} \, z_{ij}^{t,l}
\right]
\right),
\end{equation}
where $\phi_v^l$ is an MLP with residual connections and ReLU activations. \(\mathcal{N}(i)\) denotes the neighborhood of node \(v_i\), and \([\cdot]\) indicates concatenation along the feature dimension.

Although the weighting mechanism is structurally related to edge-conditioned attention, it differs in that the compatibility score in Eq. \ref{eq:anisotropic} is conditioned on the edge embedding $z_{ij}$, which is initialized exclusively from geometric features (displacement vector and Euclidean distance) and carries no semantic node-feature information at layer $l = 0$. This geometric initialization biases the learned weights toward physically motivated transport directions, in contrast to standard attention mechanisms where scores are derived from unconstrained node representations. 

\subsection{GraphConv--LSTM for Latent Temporal Dynamics}
To mitigate error accumulation over multiple rollout steps, an LSTM cell is integrated after each message-passing block, allowing the model to capture and propagate temporal dependencies in the latent node embeddings. Specifically, we follow the Graph Convolutional LSTM formulation proposed by Seo et al.~\cite{seo2018structured}, where the standard convolution in ConvLSTM is replaced by a spectral graph convolution operator \( *_G \).

Let \(\mathbf{Z}^{t,L} \in \mathbb{R}^{n \times d_v}\) denote the matrix of node embeddings obtained from the message-passing module at last step \(L\) and time t, and let \(\mathbf{H}^{t-1}\), \(\mathbf{C}^{t-1}\) be the hidden and cell states of the LSTM at the previous time-step. The GraphConv-LSTM cell updates are defined as:

\begin{equation}
\begin{aligned}
    \mathbf{i}^t &= \sigma(\mathbf{W}_{xi} *_G \mathbf{Z}^{t,L} + \mathbf{W}_{hi} *_G \mathbf{H}^{t-1} + \mathbf{b}_i), \\
    \mathbf{f}^t &= \sigma(\mathbf{W}_{xf} *_G \mathbf{Z}^{t,L} + \mathbf{W}_{hf} *_G \mathbf{H}^{t-1} + \mathbf{b}_f), \\
    \mathbf{C}^t &= \mathbf{f}^t \odot \mathbf{C}^{t-1} + \mathbf{i}^t \odot \tanh(\mathbf{W}_{xc} *_G \mathbf{Z}^{t,L} + \mathbf{W}_{hc} *_G \mathbf{H}^{t-1} + \mathbf{b}_c), \\
    \mathbf{o}^t &= \sigma(\mathbf{W}_{xo} *_G \mathbf{Z}^{t,L} + \mathbf{W}_{ho} *_G \mathbf{H}^{t-1} + \mathbf{W}_c \odot \mathbf{C}^t + \mathbf{b}_o), \\
    \mathbf{H}^t &= \mathbf{o}^t \odot \tanh(\mathbf{C}^t).
\end{aligned}
\end{equation}

Here, \(\mathbf{i}^t\), \(\mathbf{f}^t\), and \(\mathbf{o}^t\) denote the input, forget, and output gates at time \(t\), respectively; \(\odot\) is the element-wise (Hadamard) product. Each gate applies a sigmoid activation \(\sigma(\cdot)\) to its affine transform, parameterized by learnable weight matrix $\mathbf{W}_*$ and bias vector  $\ \mathbf{b}_*$. The graph convolution operator \(*_G\) is implemented using Chebyshev polynomials~\cite{defferrard2016convolutional}.

After $L$ message-passing layers, node embeddings are collected in
$\mathbf{Z}^{t} \in \mathbb{R}^{n \times d_v}$.
These embeddings are processed by a GraphConv--LSTM, which captures temporal dependencies while preserving spatial coupling through graph convolutions.
The module outputs hidden states $\mathbf{H}^{t} \in \mathbb{R}^{n \times d_v}$.

\subsection{Decoder and Time Integration}
After message passing, the final node embeddings are processed by a GraphConv-LSTM, which captures temporal dependencies directly in the latent space and produces hidden states for forecasting. In contrast to~\cite{ju2024learning}, which predicts the next state directly, we adopt a residual formulation in which the decoder predicts increments of the physical state rather than absolute values. The dynamic state is advanced using an explicit Euler update, as proposed by~\cite{eliasof2024data}:
\begin{equation}
X_{\mathrm{dyn}}^{t+1} = X_{\mathrm{dyn}}^{t} + \Delta t \; \phi_{\mathrm{dec}}(\mathbf{H}^{t}),
\qquad \mathbf{H}^{t} = \mathrm{GraphConvLSTM}(\mathbf{Z}^{t}).
\label{eq:decoder}
\end{equation}

\subsection{Training Objective}
The model is trained using autoregressive multi-step supervision over horizons of length $T$. Given simulator-generated trajectories $\{\hat{X}^t\}_{t=1}^{T}$, the loss is defined as the mean squared error between predicted and ground-truth states over all nodes and rollout steps:
\begin{equation}
\mathcal{L}(\theta) =
\frac{1}{T-T_{train}}
\sum_{t=1}^{T-T_{train}}
\frac{1}{T_{train}}
\sum_{j=1}^{T_{train}}
\frac{1}{n}
\left\lVert \hat{X}^{t+j} - X^{t+j} \right\rVert_2^2.
\end{equation}
All variables are normalized using Z-score statistics computed on the training set. During training, the rollout is executed in closed-loop mode: at each step $j$ within the supervision window, the model receives its own prediction $X^{t+j-1}$ as input rather than the ground-truth state, matching the autoregressive deployment at inference time. No teacher forcing or scheduled sampling is applied. Gradients are back-propagated through the full rollout window without truncation. This end-to-end multi-step training directly exposes the model to its own prediction errors during optimization, encouraging rollout-stable behavior and reducing the distribution shift between training and inference that arises in one-step-supervised models such as the original MeshGraphNet \cite{Pfaff2021MeshGraphNet}.

\section{Experimental Setup and Results}\label{sec:exp}

\subsection{Simulation Accuracy and Spatial--Temporal Resolution}

\begin{figure}[htpb]
    \centering
    \includegraphics[width=\linewidth]{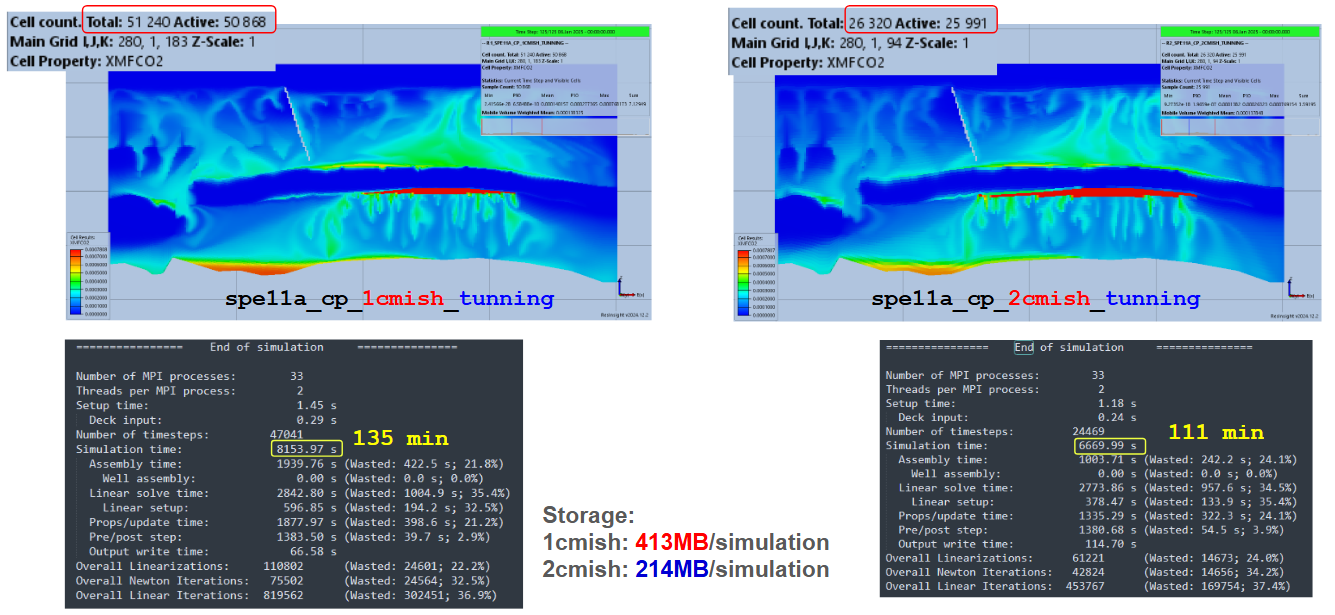}
    \caption{SPE11A grid resolution comparison.
    Left: liquid-phase density computed using the benchmark $1\,\mathrm{cm}$ grid.
    Right: liquid-phase density computed using a $2\,\mathrm{cm}$ grid.
    The figure also reports the total simulation time and storage requirements for both resolutions.}
    \label{fig:2cmish}
\end{figure}

\begin{figure}[htpb]
    \centering
    \includegraphics[width=\linewidth]{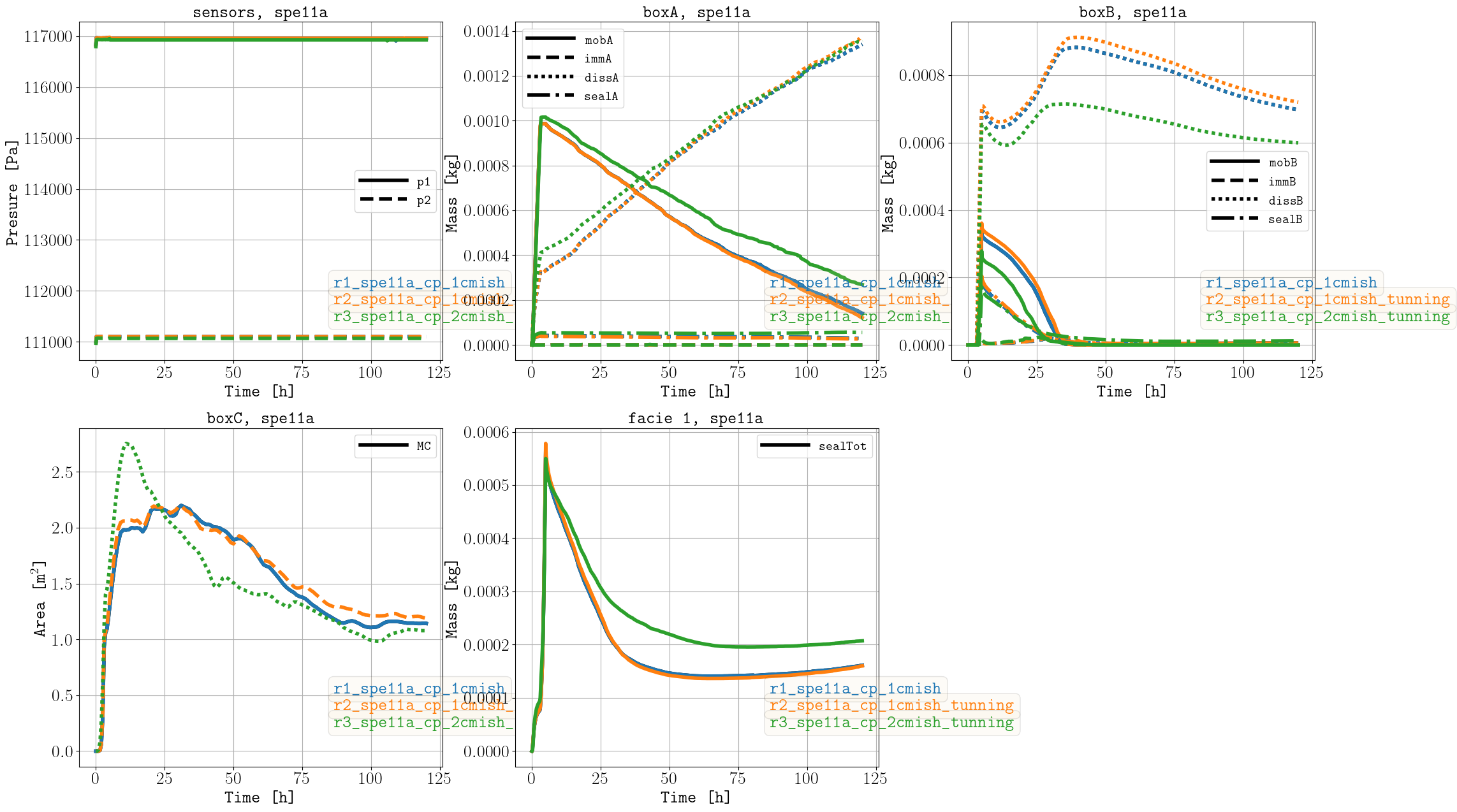}
    \caption{Comparison of SPE11A sparse quantities computed with grid resolutions of $1\,\mathrm{cm}$ and $2\,\mathrm{cm}$.
    All benchmark-reported sparse metrics exhibit good agreement across resolutions.}
    \label{fig:spe11_sparse_data}
\end{figure}

Numerical experiments are conducted using the SPE11A benchmark~\cite{nordbotten202411th}, with simulations generated through the \texttt{pyopmspe11} framework~\cite{LandaMarban2025} using corner-point grids. Figure \ref{fig:2cmish} shows, in the top-left panel, the liquid-phase density at the end of the simulation computed using the benchmark resolution, namely the 1 cm grid. In the top-right panel, the liquid-phase density is computed on the coarser grid with a 2 cm resolution. Both solutions are similar but show visible discrepancies, particularly in the finger pattern, which are better resolved with the fine grid. The panels at the bottom of Figure \ref{fig:2cmish} give the computer times for each solution (135 min for the 1 cm resolution, and 111 min for the 2 cm resolution), and the data size for each one, that is, 413MB for the 1 cm resolution, and 214MB for the 2 cm resolution. In addition to spatial maps of the field variables, Figure \ref{fig:spe11_sparse_data} shows the following measurables, details of which, including time resolution, are described in \cite{nordbotten202411th}. Top left, pressure in two points, top center and right, phase composition; Bottom left, the extent of convective mixing, and in the bottom right, the total mass of CO2 in all sealing units. In all the panels of Figure \ref{fig:spe11_sparse_data}, we plot three curves: in light blue, the benchmark solution. In orange, a tuned solution with a 1 cm resolution, and in green, the tuned 2 cm resolution solution. Comparing the data in Figures \ref{fig:2cmish} and \ref{fig:spe11_sparse_data}, we see that the 2 cm resolution solution provides a good compromise between accuracy and computational cost (processing time and storage space). All simulations from now on are performed on a corner-point grid with a spatial resolution of $2\,\mathrm{cm}$.
Although the official SPE11A benchmark adopts a $1\,\mathrm{cm}$ resolution, our results demonstrate that the coarser grid provides quantitatively consistent predictions for gas saturation, liquid-phase density, and all sparse quantities defined in the benchmark.
This choice reduces storage requirements by approximately $50\%$, enabling more efficient dataset generation and model training.
System states are sampled every $\Delta t = 90\,\mathrm{s}$, yielding trajectories of $N = 4{,}800$ time steps, which correspond to a total simulated time of five days ($120$ hours).

\subsection{Training and Testing Dataset}
\begin{figure}[htbp]
    \centering

    \begin{subfigure}[b]{0.45\textwidth}
        \centering
        \includegraphics[width=\textwidth]{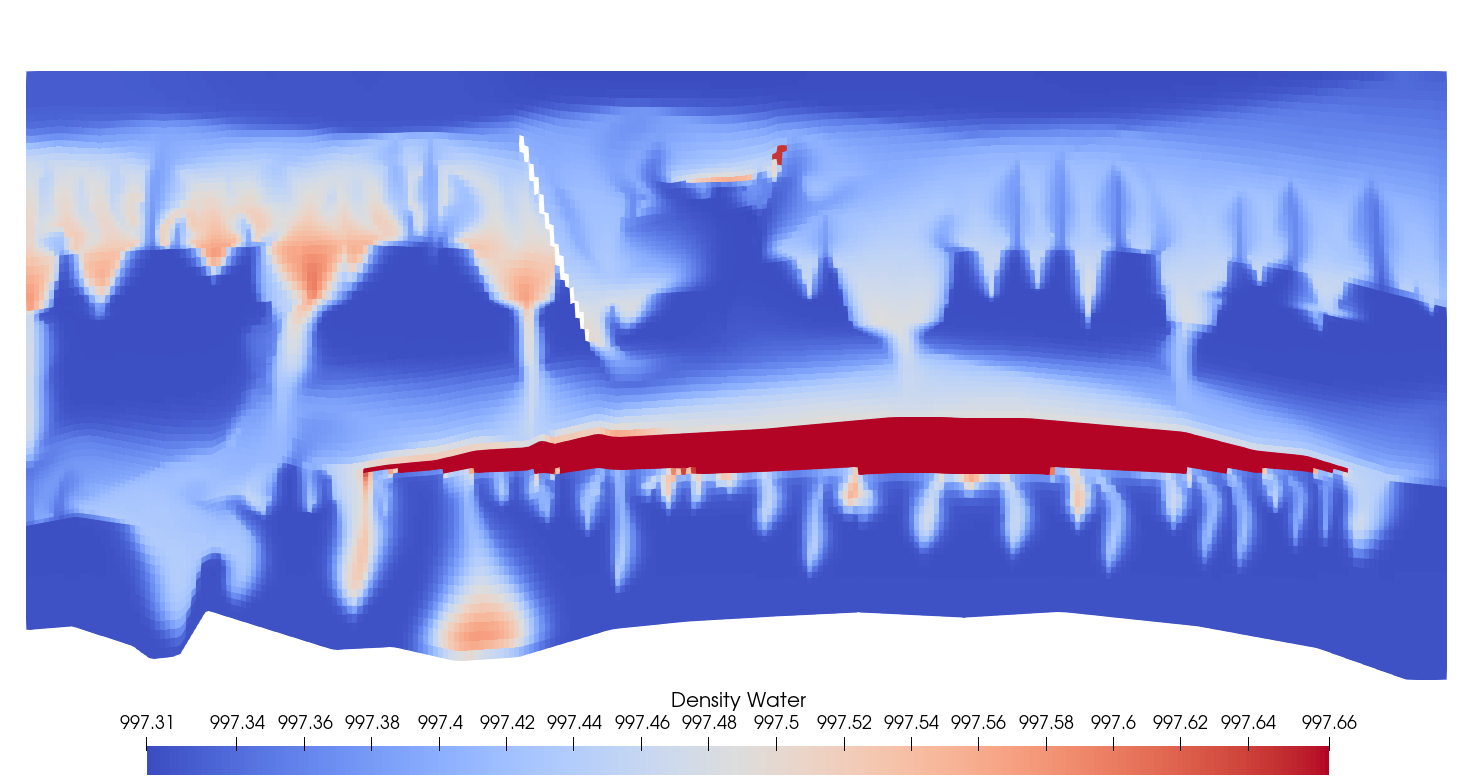}
        \caption{Trajectory 3.}
        \label{fig:traj3}
    \end{subfigure}
    \hfill
    \begin{subfigure}[b]{0.45\textwidth}
        \centering
        \includegraphics[width=\textwidth]{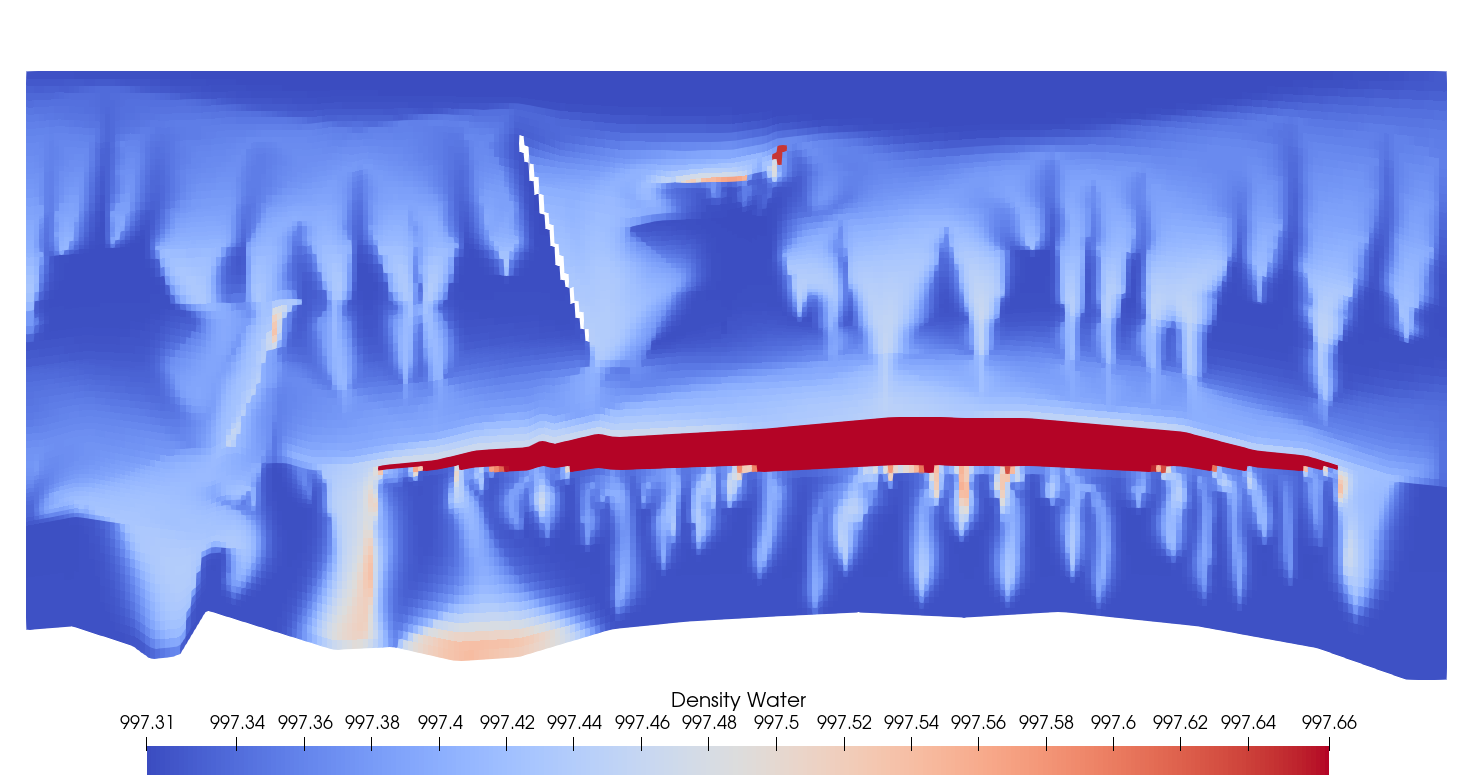}
        \caption{Trajectory 6.}
        \label{fig:traj6}
    \end{subfigure}

    \vspace{1em} 

    \begin{subfigure}[b]{0.45\textwidth}
        \centering
        \includegraphics[width=\textwidth]{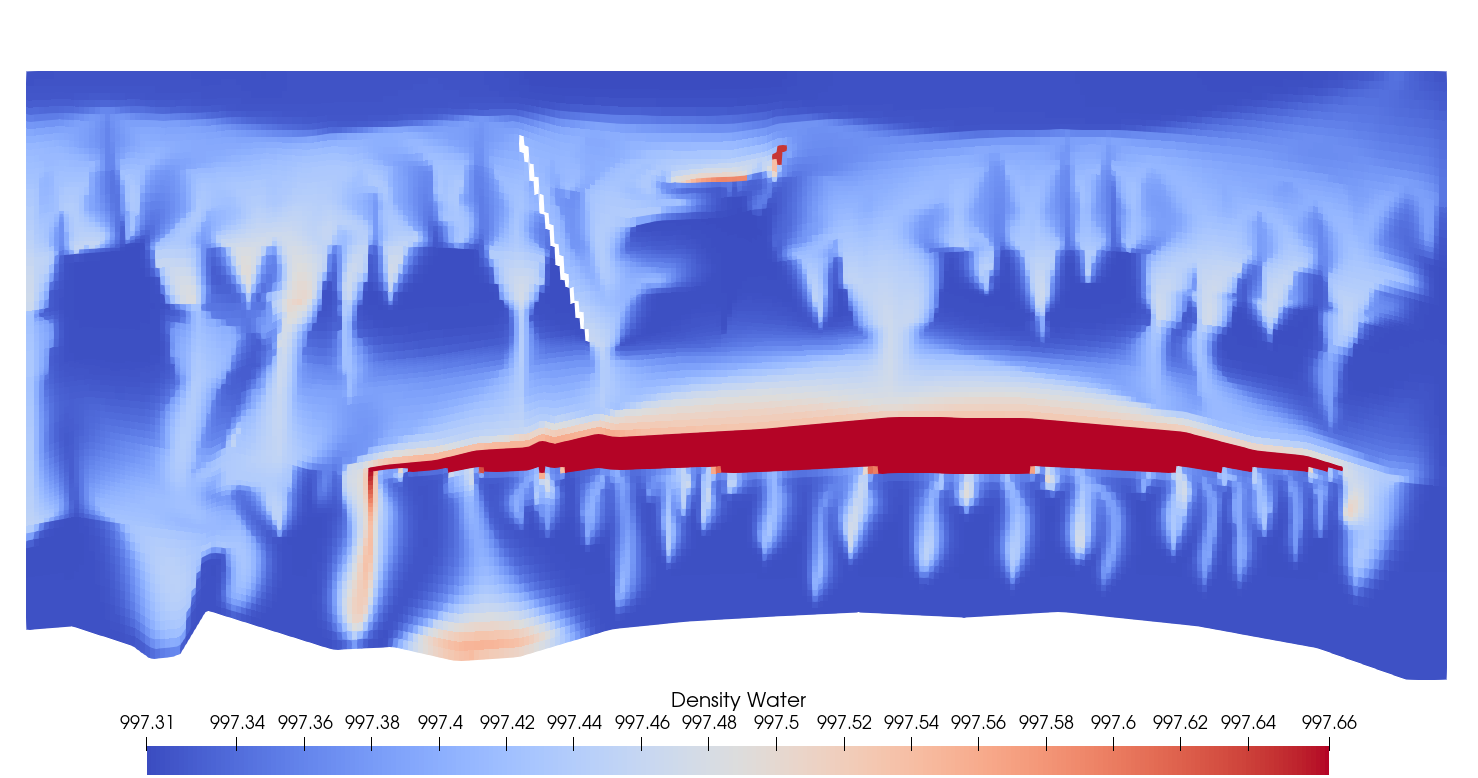}
        \caption{Trajectory 10.}
        \label{fig:traj10}
    \end{subfigure}
    \hfill
    \begin{subfigure}[b]{0.45\textwidth}
        \centering
        \includegraphics[width=\textwidth]{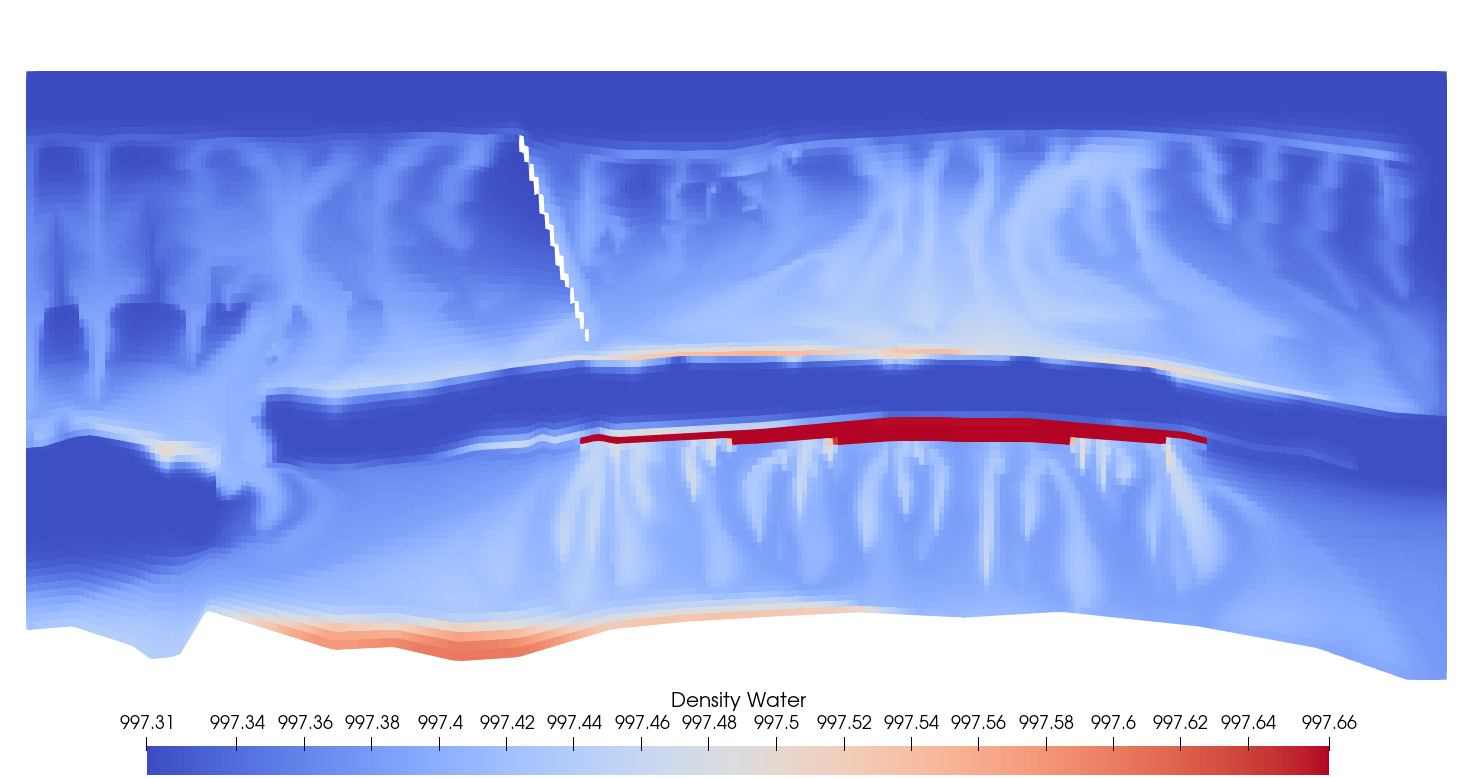}
        \caption{Ground truth.}
        \label{fig:GT}
    \end{subfigure}

    \caption{Liquid-phase density at time $t=120$ hours for   three trajectories (a)-(c), out of ten with randomly generated permeabilities and porosities for the  seven facies, and (d) the SPE11A benchmark ground truth. All results are computed with the 2 cm resolution grid.}
    \label{fig:trajectories}
\end{figure}

We then generate 10 trajectories by randomly varying the permeability and porosity of the seven facies specified in the benchmark. The purpose is to generate a rich set of complex porous media at laboratory scale, keeping the properties within the bounds of all benchmark contributors~\cite{Flemisch2024,SaloSalgado2024}. Figure \ref{fig:trajectories} shows snapshots of the liquid-phase density at $t=120$ hours for three different trajectories and the SPE11A simulation results with the 2 cm resolution grid. We can clearly identify in Figures \ref{fig:trajectories} (a) to (c) a rich variability in the liquid-phase density induced by the different facies' permeabilities and porosities. We note strikingly different fingering patterns, highlighting the richness in the generated data. Although the dataset comprises only 10 trajectories, the strikingly different fingering patterns visible in Figure 3 indicate that the geological realizations span a wide range of flow regimes, reducing the risk of overfitting to a narrow region of the parameter space. The high standard deviations observed in the 1--step RMSE entries of Table 3, where the standard deviation exceeds the mean for several injection-regime cases, reflect the strong variability in early-time CO2 spreading across realizations, rather than model instability.

The trajectory set is sampled every $\Delta t = 90\,\mathrm{s}$, resulting in trajectories of length $N = 4800$ time steps, corresponding to $5$ days of physical time. Generating all 10 trajectories takes approximately 3.7 days on an Intel Xeon Gold. The SPE11A benchmark exhibits two distinct dynamical regimes: an initial injection phase in which $CO_2$ is injected at two wells, followed by a longer post-injection phase with zero injection. In our setup, injection is active until 200 time steps ($t=18\times10^3$s), while the interval $t \in (18\times 10^3,432\times10^3]$s (time steps 201--4800) corresponds to a no-injection regime, leading to a strong imbalance between the injection and no-injection regimes.

To assess the impact of this imbalance on long-horizon forecasting, we train and evaluate three models: (i) a model trained predominantly on the injection regime, $t \leq 27\times10^3$s (300 time steps), (ii) a model trained only on the post-injection regime, $t > 27\times10^3$s (time steps 301 to 4800), and (iii) a model trained on the whole trajectory spanning both regimes. This experimental design allows us to evaluate the robustness of the proposed architecture to regime shifts and data imbalance.

We used a 90/10 train-validation split with data from 10 trajectories. We evaluated the model on a trajectory generated from an unseen geological parameter configuration using the same Cartesian grid for the flow simulation. The prediction targets are the gas saturation and liquid-phase density fields, which provide complementary information about $CO_2$ plume evolution and the density-driven fingering in the predominantly water-filled domain.

\subsection{Data Preparation and Graph Construction}
\begin{figure}[htbp]
    \centering
   
    \begin{subfigure}[b]{0.45\textwidth}
        \centering
        \includegraphics[width=\textwidth]{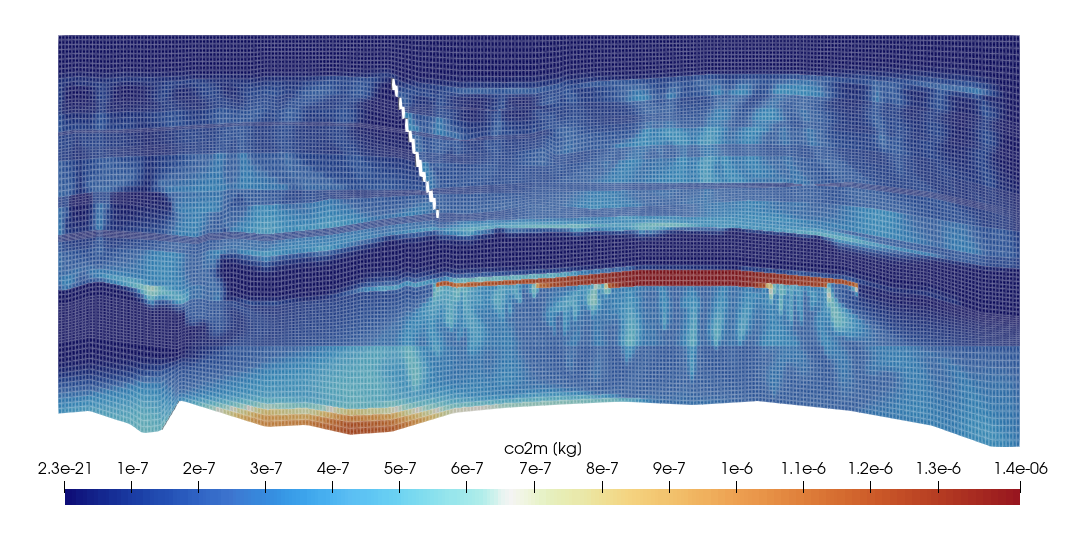}
        \caption{Mesh.}
        \label{fig:mesh}
    \end{subfigure}
    \hfill
    \begin{subfigure}[b]{0.45\textwidth}
        \centering
        \includegraphics[width=\textwidth]{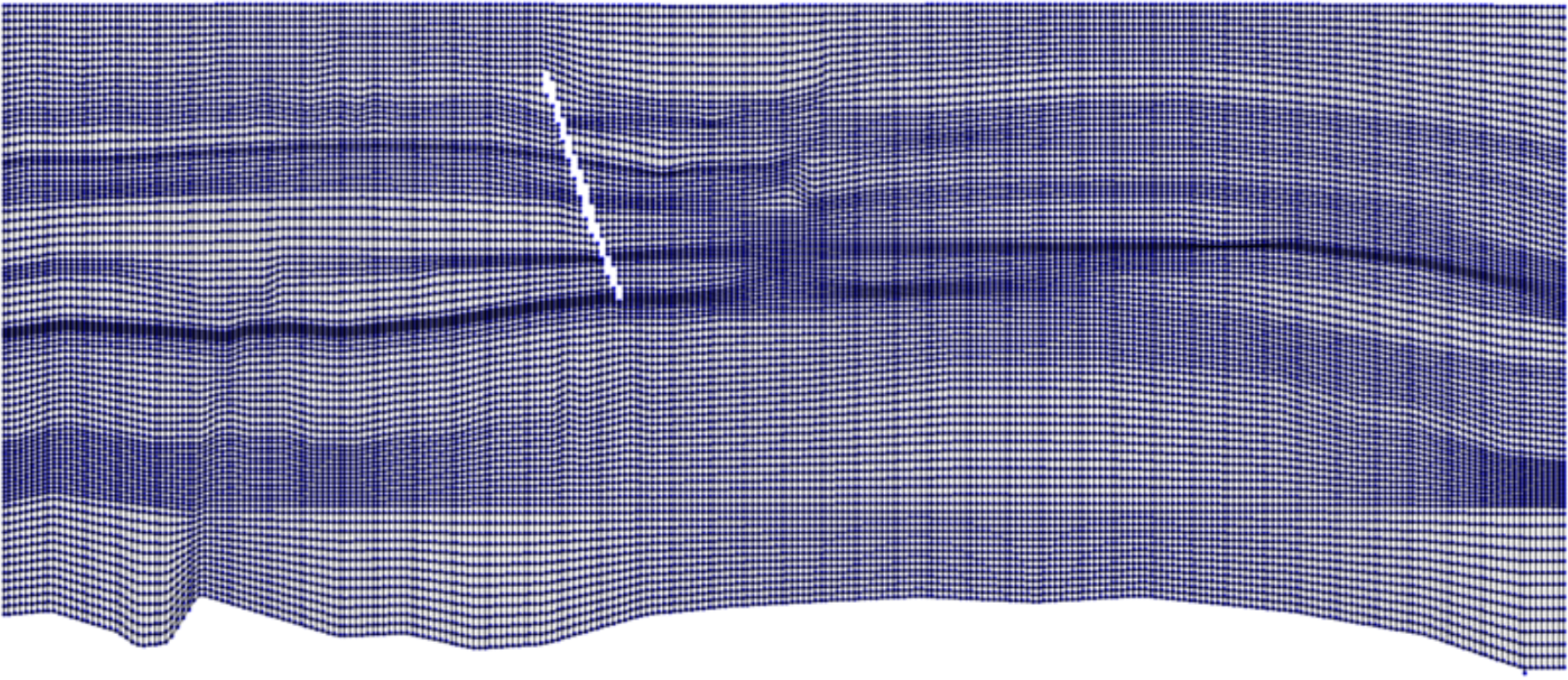}
        \caption{Graph.}
        \label{fig:graph}
    \end{subfigure}


    \caption{Left: Mesh for the SPE11A benchmark with 2 cm resolution colored by the CO2 mass at $t=120$ hours; Right: Graph, with vertices at cell barycenters and edges linking the graph vertices (compatible with transmissibility, which are associated to the face edges).}
    \label{fig:mesh-graph}
\end{figure}

To be used as input for the GNN, the SPE11A mesh, composed of hexahedral cells, is first extruded to a two-dimensional mesh (see Figure \ref{fig:mesh}), which is converted to the graph shown in Figure \ref{fig:graph}, where graph nodes correspond to two-dimensional mesh cell barycenters and graph edges represent the connection between adjacent cells (faces in two dimensions).

The static node features are cell volume, isotropic scalar permeability, porosity, cell type (well, fault, boundary, or normal), and cell-center coordinates. In contrast, the static edge features are the displacement vector and the Euclidean distance. Moreover, the time-dependent dynamic node features include the injection rate (present only in well nodes) and the current values of the target variables, i.e., gas saturation and liquid-phase density.

The computational mesh is represented as a static graph whose topology remains fixed throughout the simulation.
Each grid cell corresponds to a node, and edges connect neighboring cells according to mesh connectivity.
The resulting graph contains $n = 25{,}992$ nodes and $m = 51{,}557$ edges.
Node features encode physical state variables and static cell properties, while edge features represent relative geometric information, including displacement vectors and Euclidean distances.

\subsection{Graph Features}
Table~\ref{tab:features_spe11} summarizes the node features, edge features, and output variables used in the SPE11A graph representation.

The cell types are divided into four mutually exclusive categories, namely \emph{well}, \emph{fault}, \emph{boundary}, and \emph{normal}, represented as a one-hot encoded vector. \emph{Well} cells correspond to injection well locations; \emph{fault} cells are located around the fault, where transmissibility is zero; \emph{boundary} cells lie on the mesh boundaries; and \emph{normal} cells comprise all remaining cells that do not fit into any of the aforementioned categories. 

\begin{table}[ht!]
\centering
\caption{Graph features for each cell $i$, each pair of cells $i,j$, and the node output of the SPE11A model (gas saturation and liquid phase density). Note that, as the meshes are extruded, three dimensions are considered for the cell-center coordinates and displacement vectors, though one coordinate will be the same for all cells. Both the output and current state features correspond to gas saturation and liquid-phase density, and therefore they have dimension 2.}
\vspace{8pt}
\begin{tabular}{llll}
 & \textbf{Feature} & \textbf{Description} & \textbf{Dimension} \\
\hline
\multirow{6}{*}{Node Features}
  & $s^t_i$ & Current state & 2 \\
  & $V_i$ & Cell volume & 1 \\
  & $k_i$ & Isotropic scalar permeability & 1 \\
  & $p_i$ & Porosity & 1 \\
  & $n_i$ & Cell type & 4 \\
  & $c_i$ & Cell-center coordinates & 3 \\
  & $I^t_i$ & Time-dependent Injection rate at well's nodes & 1 \\
  \hline
\multirow{2}{*}{Edge Features}
  & $c_i - c_j$ & Displacement vector & 3 \\
  & $||c_i - c_j||$ & Euclidean distance & 1 \\
  \hline
\multirow{1}{*}{Output}
  & $s^{t+1}_i$ & Next step state & 2 \\
\end{tabular}
\label{tab:features_spe11}
\end{table}

\subsection{Training Setup and Hyperparameters}\label{sec:training_hyperparameters}
Table~\ref{tab:hyperparameters} summarizes the architectural and training hyperparameters used in all experiments.
Unless otherwise stated, the same configuration is adopted across all simulation trajectories and prediction targets.

The model is trained using multi-step supervision  with the Adam optimizer.
Multi-step forecasts are obtained via autoregressive rollouts at inference time. The hyperparameters in Table~\ref{tab:hyperparameters} have been chosen after extensive ablation studies. In particular, the number of message-passing layers $L = 20$, although quite large relative to the graph diameter of a 2D corner-point grid, follows the guidelines in \cite{tesan2026under}.  The Chebyshev order $K = 8$ was selected to cover the spectral bandwidth expected from the heterogeneous permeability field, and was confirmed via ablation.

\begin{table}[ht!]
\centering
\caption{Model and training hyperparameters used in all experiments.}
\label{tab:hyperparameters}
\vspace{5pt}
\begin{tabular}{l l}
\textbf{Hyperparameter} & \textbf{Value} \\
\hline
Latent node dimension ($d_v$) & 32 \\
Latent edge dimension ($d_e$) & 32 \\
MLP hidden size & 128 \\
Number of message-passing layers ($L$) & 20 \\
GraphConv--LSTM hidden size & 32 \\
Chebyshev order ($K$) & 8 \\
Optimizer & Adam \\
Learning rate & $1 \times 10^{-3}$ \\
Batch size & 8 \\
Training epochs & 200 \\
Time step ($\Delta t$) & $90\,\mathrm{s}$ \\
Multi-step prediction (T) & $10$\\
\end{tabular}
\end{table}

\subsection{Results}
To assess the model's predictive capability, gas saturation and liquid-phase density are forecast over 1-, 10-, and 50-step autoregressive rollouts initialized at different time instances. The ground truth consists of simulator outputs generated from an unseen geological configuration whose facies permeabilities and porosities match the SPE11A benchmark specification.

\begin{table}[htpb!]
\centering
\caption{
Cumulative RMSE under different training regimes and forecasting horizons.
Errors are reported for gas saturation ($S_g$, dimensionless) and liquid-phase density ($\rho_\ell$, kg\,m$^{-3}$).
All table entries are scaled by $\times 10^{-3}$.
}
\label{tab:cumulative_rmse_models_inverted}
\setlength{\tabcolsep}{4pt}
\renewcommand{\arraystretch}{1.05}
\vspace{5pt}
\begin{tabular}{lcccccc}
\textbf{Forecast Horizon}
& \multicolumn{2}{c}{\textbf{Injection regime only}}
& \multicolumn{2}{c}{\textbf{Post-injection regime}}
& \multicolumn{2}{c}{\textbf{Both regimes}} \\
\cline{2-3}\cline{4-5}\cline{6-7}
& \textbf{$S_g$} & \textbf{$\rho_\ell$}
& \textbf{$S_g$} & \textbf{$\rho_\ell$}
& \textbf{$S_g$} & \textbf{$\rho_\ell$} \\
\hline
RMSE (1-step)   &\tiny $1.164 \pm 1.064$    & \tiny $0.793 \pm 0.920$   & \tiny  $0.553 \pm 1.190$ & \tiny  {$0.42 \pm 0.82$} & \tiny $0.562 \pm 1.152 $  & \tiny  $0.476 \pm 0.963$ \\
RMSE (10-steps) & \tiny $15.22 \pm 4.75$   & \tiny$10.743 \pm 2.86$  & \tiny {$4.80 \pm 4.71$} & \tiny  $3.247 \pm 2.67$ & \tiny $4.946 \pm 4.622$  & \tiny $3.434\pm 2.965$ \\
RMSE (50-steps) & \tiny $85.35 \pm 21.41$ & \tiny $71.47 \pm 10.34$ & \tiny $23.91 \pm 10.44$ & \tiny $16.40 \pm 6.263$ & \tiny $24.734 \pm 10.673$ & \tiny $16.914 \pm 6.457$ \\
\hline
\end{tabular}
\label{tab:rmse}
\end{table}

Table~\ref{tab:rmse} reports the cumulative RMSE as a function of the rollout horizon. As expected for autoregressive graph-based simulators, the prediction error increases with the forecasting horizon due to temporal error propagation. However, the error growth remains bounded over the evaluated rollout horizons, including extended 50-step rollouts. Crucially, the magnitude of the errors remains relative to the physical scale of the state variables.

For liquid-phase density, whose typical values lie between $997$ and $998\,\text{kg}\,\text{m}^{-3}$, the 50-step cumulative RMSE of the joint model corresponds to an absolute deviation of approximately $1.7 \times 10^{-2}\,\text{kg}\,\text{m}^{-3}$ (after accounting for the $\times 10^{-3}$ scaling), yielding a relative error on the order of $10^{-5}$. This level of accuracy indicates that the learned simulator preserves the fidelity of the density field over long autoregressive horizons.

For gas saturation, which is bounded between 0 and 1, the 50-step cumulative RMSE of the joint model remains below $2.5\%$, suggesting that the model captures the main plume migration dynamics without evident instability in the evaluated rollouts. For several 1-step rollout cases, the standard deviation exceeds the mean RMSE, likely reflecting the high variability inherent in the early injection regime. Notably, this performance is achieved under unseen geological configurations, suggesting that the proposed \emph{AnisoMeshGraph-LSTM} generalizes to such configurations within the tested setup while error remains well controlled. 
  
Training on all the data in the 10 trajectories takes 22 hours on an NVIDIA H100 with 94GB of RAM, and inference for 50 steps takes only 160 seconds on the same machine. Recalling that the time for a single simulation is between 111 and 135 minutes (see Figure \ref{fig:2cmish}), we see that forecasting gas saturation and liquid-phase density at 50-step intervals using the present GNN model is very fast.   

\begin{figure}[!htb]
    \centering

    \begin{subfigure}{\linewidth}
        \centering
        \includegraphics[width=\linewidth]{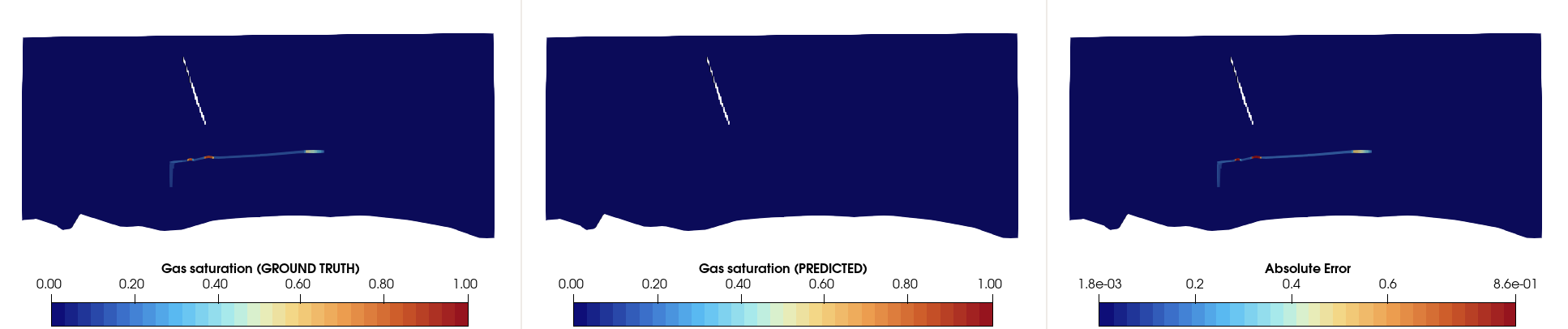}
    \end{subfigure}

    \begin{subfigure}{\linewidth}
        \centering
        \includegraphics[width=\linewidth]{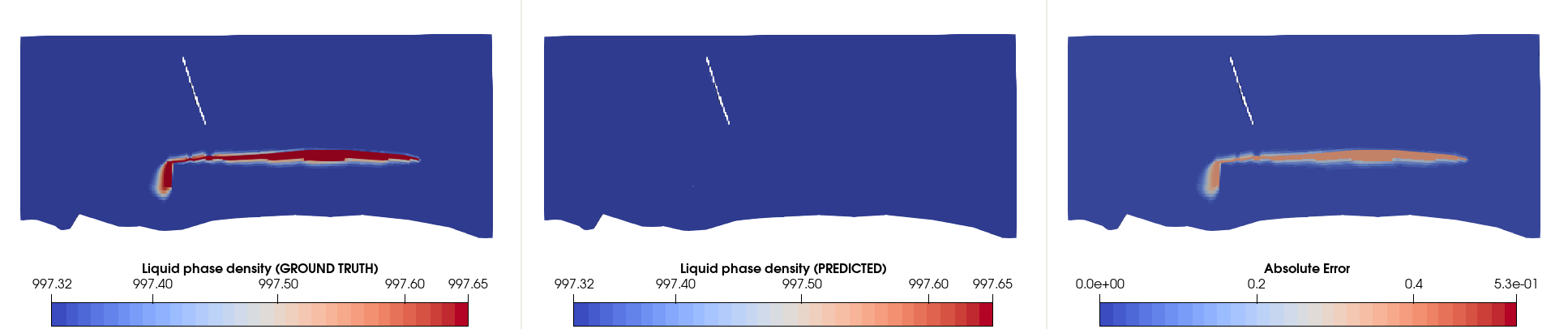}
    \end{subfigure}

    \caption{Prediction at time $t=900$s (starting inference from $t=0$s) using the model trained during both regimes. Top, gas saturation; bottom, liquid phase density ($kg\,m^{-3}$). Note the large absolute errors at early injection time, reflecting the rapid and difficult-to-predict initial CO2 spreading.}
    \label{fig:c1}
\end{figure}

\begin{figure}[!htb]
    \centering

    \begin{subfigure}{\linewidth}
        \centering
        \includegraphics[width=\linewidth]{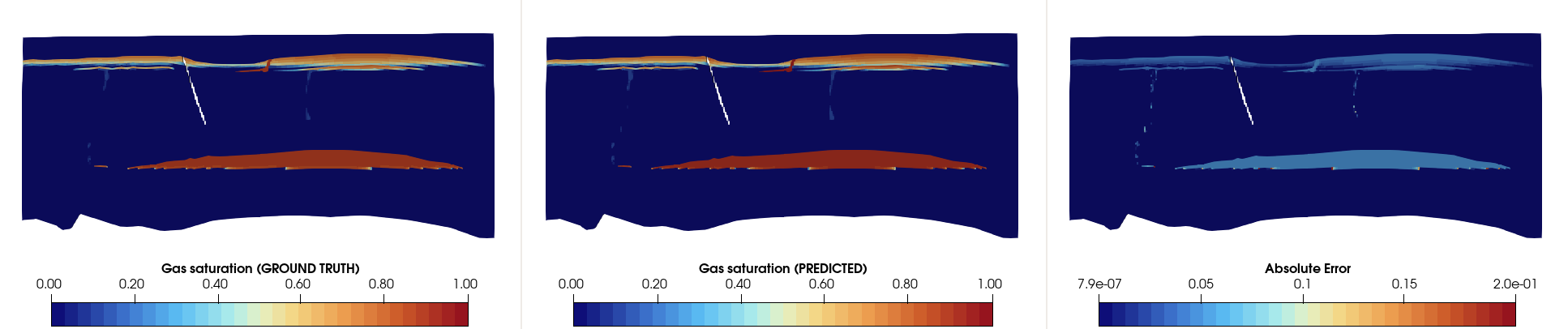}
    \end{subfigure}

    \begin{subfigure}{\linewidth}
        \centering
        \includegraphics[width=\linewidth]{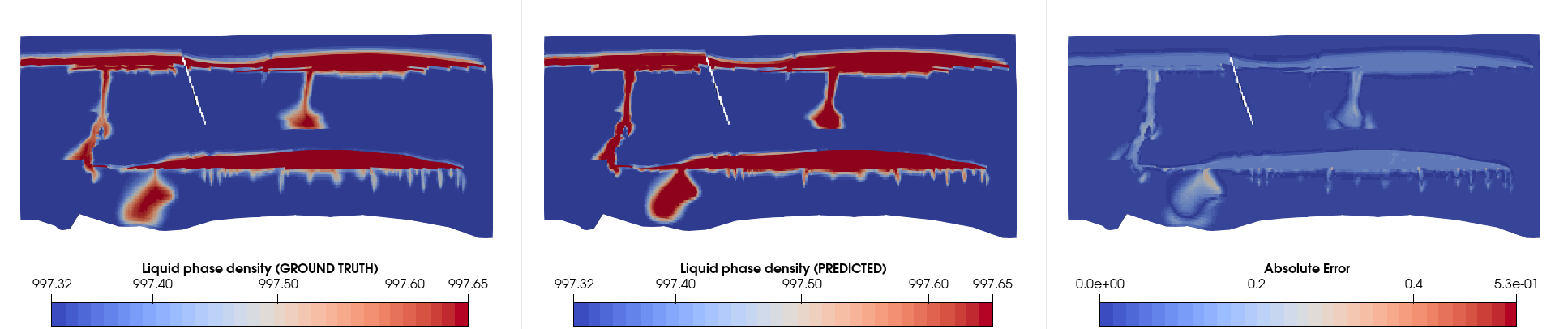}
    \end{subfigure}

    \caption{Prediction at time $t=23{,}400$s (starting inference from $t=22{,}500$s) using the model trained during both regimes. Top, gas saturation; bottom, liquid phase density ($kg\,m^{-3}$). The absolute error is spatially localized at plume fronts and finger tips, where nonlinear advection dominates.}
    \label{fig:c3}
\end{figure}

\begin{figure}[!htb]
    \centering
    \includegraphics[width=0.65\linewidth]{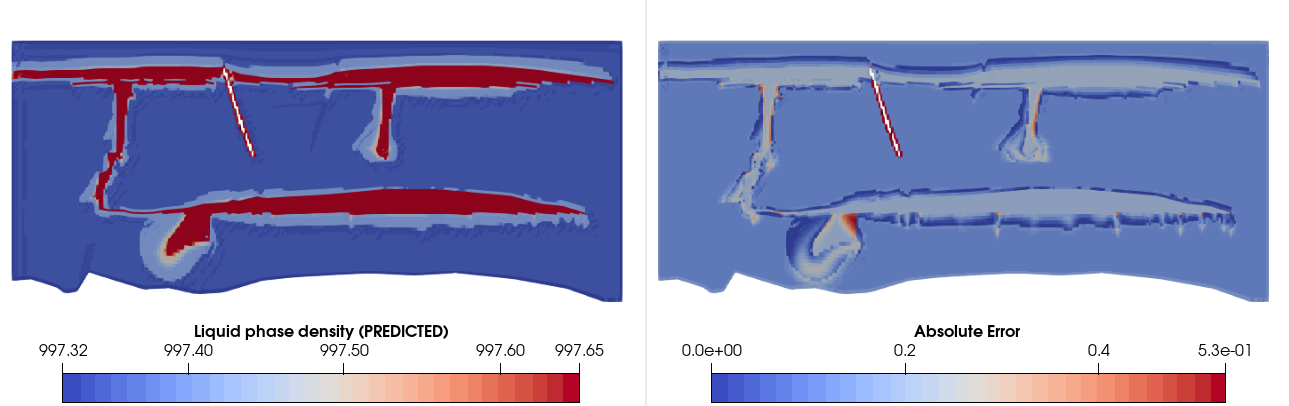}
    \caption{Liquid-density prediction and absolute error produced by the baseline model at time $t=23,400$ seconds. Compared with Figure \ref{fig:c3} (bottom row), the baseline produces substantially larger and more diffuse errors, particularly in the fingering region. Note that both figures share the same colorbar scale for the absolute error, allowing direct visual comparison of error magnitude and spatial distribution between the two models.}
    \label{fig:baseline}
\end{figure}

Figure \ref{fig:c1} shows a representative 10-step-ahead prediction at the early stages of the simulation, at the start of injection using the model trained during both injection regimes. The early stages are a challenging period for CO$_2$ plume prediction, since the gas spreads significantly faster than in other stages of the simulation. As we can see, the errors are relatively large, and the model's predictions are not good. However, for later times, predictions improve considerably. Figure~\ref{fig:c3} shows a representative 10-step-ahead prediction with $\Delta t = 90\,\mathrm{s}$ in the post-injection regime, forecasting the system state at $t = 23{,}400\,\mathrm{s}$ from an initialization at $t = 22{,}500\,\mathrm{s}$. The model reproduces the dominant transport features governed by the underlying multiphase flow equations, including sharp gas--water interfaces and the complex density-driven fingering patterns that emerge after the end of CO$_2$ injection. Both gas saturation and liquid-phase density fields exhibit good qualitative agreement with the ground truth, while the absolute error remains spatially localized near plume fronts and finger tips, which correspond to regions dominated by strong nonlinear advection. This behavior is consistent with the cumulative RMSE trends reported in Table~\ref{tab:rmse}, indicating that the anisotropic message-passing GNN captures directional transport effects over extended horizons.
Such good predictions are not observed for a model trained without considering the anisotropic message-passing introduced in the present paper and without updating the next state directly, as in the model used in  \cite{ju2024learning}, which we identify as the baseline model. Figure \ref{fig:baseline} shows the liquid-density prediction and absolute error produced by the baseline model at time $T=23,400$ seconds. Comparing the results in this figure with the prediction and absolute error produced by the present model in the bottom row of Figure~\ref{fig:c3}, we can clearly see, by inspecting the hot colors in the absolute error's colorbar, that the prediction obtained with the present model is better and has lower error than the baseline model prediction. Figure \ref{fig:c7} shows the later stages of the simulation, when dissolution dominates. In these stages, the model predictions remain in close agreement with the ground truth values, representing with good accuracy the complex fingering patterns.

\begin{figure}[!htb]
    \centering

    \begin{subfigure}{\linewidth}
        \centering
        \includegraphics[width=\linewidth]{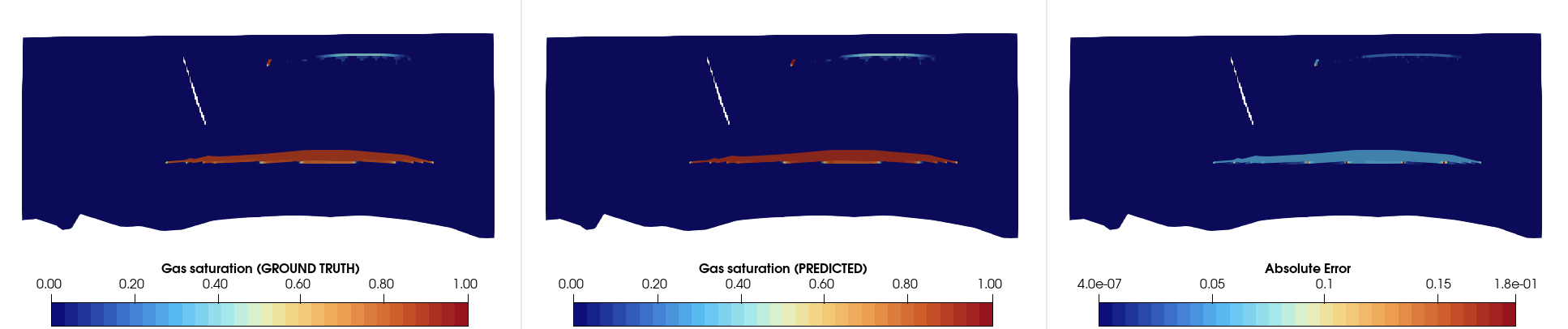}
    \end{subfigure}

    \begin{subfigure}{\linewidth}
        \centering
        \includegraphics[width=\linewidth]{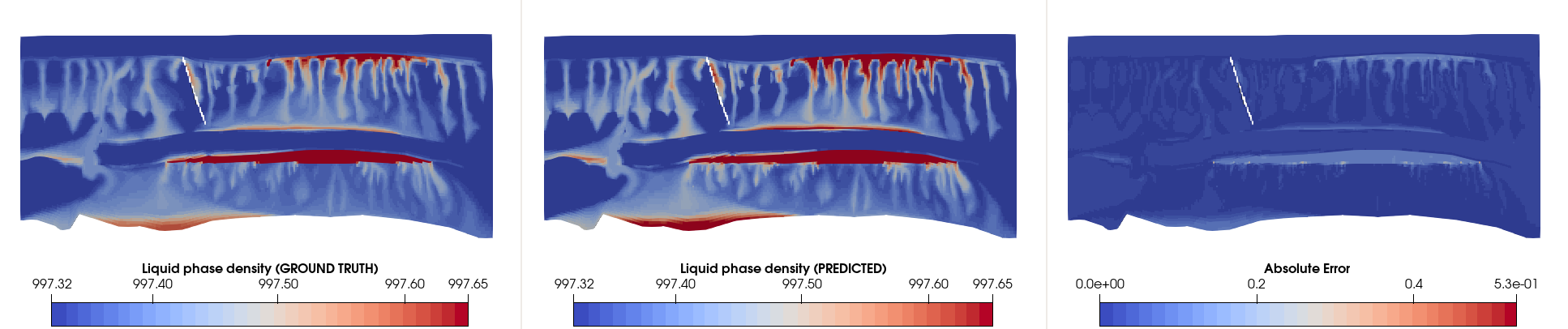}
    \end{subfigure}

    \caption{Prediction at time $t=225{,}900$s (starting inference from $t=225{,}000$s) using the model trained during both regimes. Top, gas saturation; bottom, liquid phase density ($kg\,m^{-3}$). The model captures the complex fingering structure with good fidelity in the intermediate dissolution regime.}
    \label{fig:c5}
\end{figure}

\begin{figure}[!htb]
    \centering

    \begin{subfigure}{\linewidth}
        \centering
        \includegraphics[width=\linewidth]{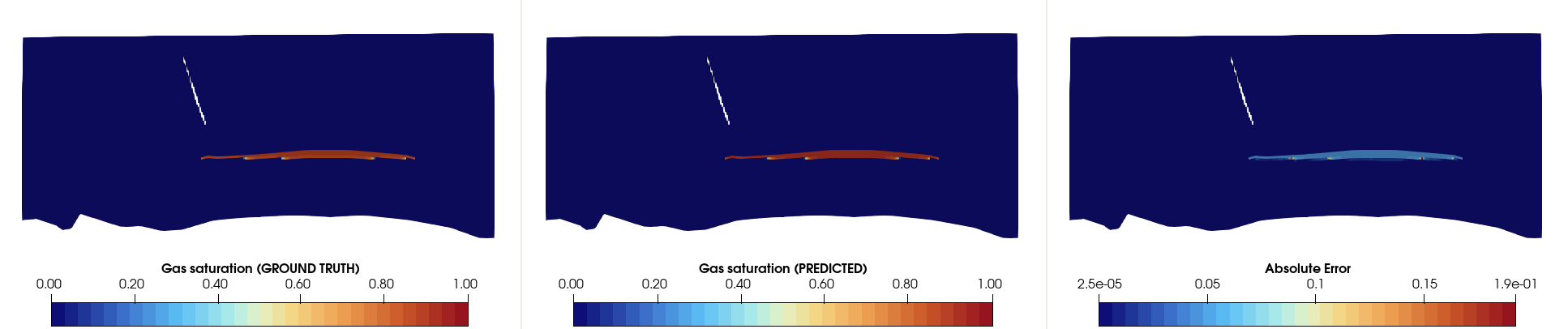}
    \end{subfigure}

    \begin{subfigure}{\linewidth}
        \centering
        \includegraphics[width=\linewidth]{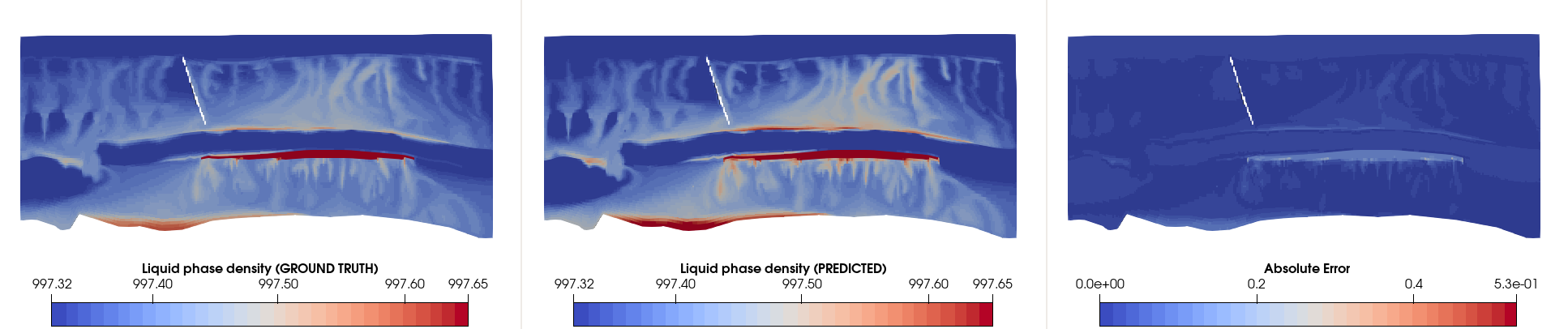}
    \end{subfigure}

    \caption{Prediction at time $t=405{,}900$s (starting inference from $t=405{,}000$s), using the model trained during both regimes. Top, gas saturation; bottom, liquid phase density ($kg\,m^{-3}$). At late times, when dissolution dominates, predictions remain in close agreement with the ground truth across the full domain.}
    \label{fig:c7}
\end{figure}

A physics-aware evaluation metric, following the approach of \cite{ju2024learning}, counts the number of cells with gas saturation below a given threshold, providing a physically interpretable measure of plume extent that complements the RMSE results above. We choose, as in \cite{ju2024learning}, the threshold value for the saturation $s=0.01$. For the following times we measure it on the ground truth (GT) and on the prediction (PR) cases, respectively: time $23,400s$ - Figure~\ref{fig:c3} - 23,666 (GT) and 23,452 (PR); time $225,900s$ - Figure~\ref{fig:c5} - 25,364 (GT) and 25,187 (PR); time $405,900s$ - Figure~\ref{fig:c7} - 25,718 (GT) and 25,577 (PR). These counts suggest a close match between ground truth and prediction for the small saturation interval, where most of the cells are present.

\section{Conclusions}\label{sec:conc}
We presented \emph{AnisoMeshGraph-LSTM}, whose central contribution is a geometry-conditioned anisotropic message-passing mechanism, combined with GraphConv--LSTM latent dynamics and a residual autoregressive formulation, for forecasting multiphase flow in geological CO$_2$ storage. By incorporating geometry-informed anisotropic message passing, the proposed model captures the strongly directional transport mechanisms that govern plume migration, including sharp gas--water interfaces and density-driven fingering. Results on the SPE11A benchmark indicate promising long-horizon forecasts under both injection and post-injection regimes, with low error growth and good generalization to unseen geological configurations. These results indicate that anisotropic spatial modeling may be beneficial in graph-based simulators and open promising directions for physically informed surrogate models in subsurface flow applications. However, the present GNN engine has limitations. Prediction accuracy at early simulation times is clearly poor. No geomechanics is considered in the present paper; the SPE11A benchmark, although complex, is a two-dimensional problem at the laboratory scale. Thus, tackling a full three-dimensional problem while considering a realistic domain with coupled flow and geomechanics remains open. Furthermore, forecasting longer rollouts remains challenging due to error accumulation. 

\subsubsection*{Author Contributions}
All authors contributed equally to the design and implementation of the research, to the analysis of the
results and to the writing of the manuscript.

\subsection*{Conflict of Interest Statement}
The authors have no conflicts of interest to declare that are relevant to the content of this article.

\subsubsection*{Acknowledgments}
This study was partially financed by the Coordenação de Aperfeiçoamento de Pessoal de Nível Superior-Brasil (CAPES)---Finance Code 001. It is also partially supported by CNPq, Brazilian Petroleum Agency, and TotalEnergies E\&P Brazil under the 1\% ANP obligation (GNN CO$_2$ Project RD24-70).

\subsubsection*{Publication Note}

This manuscript has been accepted for publication as a chapter in the forthcoming book
\textit{Scientific Machine Learning for Predictive Modeling: Bridging Data-Driven and Physics-Based Approaches in Computational Science and Engineering},
edited by A. Cunha Jr., F. P. Santos, F. A. Rochinha, and A. L. G. A. Coutinho, to be published by Springer Nature.
The final authenticated version will be available through Springer Nature.

\bibliographystyle{spmpsci}
\bibliography{references}

@inproceedings{thurlemann2023anisotropic,
  title={Anisotropic message passing: Graph neural networks with directional and long-range interactions},
  author={Th{\"u}rlemann, Moritz and Riniker, Sereina},
  booktitle={The Eleventh International Conference on Learning Representations},
  year={2023}
}

@article{ju2024learning,
title = {Learning {CO2} plume migration in faulted reservoirs with Graph Neural Networks},
journal = {Computers \& Geosciences},
volume = {193},
pages = {105711},
year = {2024},
issn = {0098-3004},
doi = {10.1016/j.cageo.2024.105711},
url = {https://www.sciencedirect.com/science/article/pii/S0098300424001948},
author = {Xin Ju and Fran{\c{c}}ois P. Hamon and Gege Wen and Rayan Kanfar and Mauricio Araya-Polo and Hamdi A. Tchelepi}
}

@inproceedings{eliasof2024data,
  title={Data-Driven Higher Order Differential Equations Inspired Graph Neural Networks},
  author={Eliasof, Moshe and Haber, Eldad and Treister, Eran and Sch{\"o}nlieb, Carola-Bibiane},
  booktitle={ICLR 2024 Workshop on AI4DifferentialEquations In Science},
  year={2024},
  url={https://openreview.net/forum?id=rJReXWFByt}
}

@article{tesan2026under,
  title={On the under-reaching phenomenon in message passing neural {PDE} solvers: Revisiting the {CFL} condition},
  author={Tes{\'a}n, Lucas and Iparraguirre, Mikel M and Gonz{\'a}lez, David and Martins, Pedro and Cueto, El{\'\i}as},
  journal={Computer Methods in Applied Mechanics and Engineering},
  volume={449},
  pages={118476},
  year={2026},
  publisher={Elsevier},
  url={https://doi.org/10.1016/j.cma.2025.118476}
}

@inproceedings{Pfaff2021MeshGraphNet,
  title        = {Learning Mesh-Based Simulation with Graph Networks},
  author       = {Pfaff, Tobias and Fortunato, Meire and Sanchez-Gonzalez, Alvaro and Battaglia, Peter W.},
  booktitle    = {International Conference on Learning Representations (ICLR)},
  year         = {2021},
  url          = {https://arxiv.org/abs/2010.03409}
}

@inproceedings{seo2018structured,
  title={Structured sequence modeling with graph convolutional recurrent networks},
  author={Seo, Youngjoo and Defferrard, Micha{\"e}l and Vandergheynst, Pierre and Bresson, Xavier},
  booktitle={International conference on neural information processing},
  pages={362--373},
  year={2018},
  organization={Springer}
}

@article{defferrard2016convolutional,
  title={Convolutional neural networks on graphs with fast localized spectral filtering},
  author={Defferrard, Micha{\"e}l and Bresson, Xavier and Vandergheynst, Pierre},
  journal={Advances in neural information processing systems},
  volume={29},
  year={2016}
}

@article{LandaMarban2025,
  doi = {10.21105/joss.07357},
  url = {https://doi.org/10.21105/joss.07357},
  year = {2025},
  publisher = {The Open Journal},
  volume = {10},
  number = {105},
  pages = {7357},
  author = {Landa-Marb{\'a}n, David and Sandve, Tor H.},
  title = {pyopmspe11: A {Python} framework using {OPM} Flow for the {SPE11} benchmark project},
  journal = {Journal of Open Source Software}
}

@Article{SaloSalgado2024,
  author={Sal{\'o}-Salgado, Llu{\'i}s and Haugen, Malin and Eikehaug, Kristoffer and Fern{\o}, Martin and Nordbotten, Jan M. and Juanes, Ruben},
  title={Direct Comparison of Numerical Simulations and Experiments of {CO2} Injection and Migration in Geologic Media: Value of Local Data and Forecasting Capability},
  journal={Transport in Porous Media},
  year={2024},
  month={Mar},
  volume={151},
  number={5},
  pages={1199--1240},
  issn={1573-1634},
  doi={10.1007/s11242-023-01972-y},
  url={https://doi.org/10.1007/s11242-023-01972-y}
}

@Article{Flemisch2024,
  author={Flemisch, Bernd and Nordbotten, Jan M. and Fern{\o}, Martin and Juanes, Ruben and Both, Jakub W. and Class, Holger and Delshad, Mojdeh and Doster, Florian and Ennis-King, Jonathan and Franc, Jacques and Geiger, Sebastian and Gl{\"a}ser, Dennis and Green, Christopher and Gunning, James and Hajibeygi, Hadi and Jackson, Samuel J. and Jammoul, Mohamad and Karra, Satish and Li, Jiawei and Matth{\"a}i, Stephan K. and Miller, Terry and Shao, Qi and Spurin, Catherine and Stauffer, Philip and Tchelepi, Hamdi and Tian, Xiaoming and Viswanathan, Hari and Voskov, Denis and Wang, Yuhang and Wapperom, Michiel and Wheeler, Mary F. and Wilkins, Andrew and Youssef, AbdAllah A. and Zhang, Ziliang},
  title={The {FluidFlower} Validation Benchmark Study for the Storage of {CO2}},
  journal={Transport in Porous Media},
  year={2024},
  month={Mar},
  volume={151},
  number={5},
  pages={865--912},
  issn={1573-1634},
  doi={10.1007/s11242-023-01977-7},
  url={https://doi.org/10.1007/s11242-023-01977-7}
}

@article{RASMUSSEN2021159,
  title = {The Open Porous Media Flow reservoir simulator},
  journal = {Computers \& Mathematics with Applications},
  volume = {81},
  pages = {159--185},
  year = {2021},
  note = {Development and Application of Open-source Software for Problems with Numerical PDEs},
  issn = {0898-1221},
  doi = {10.1016/j.camwa.2020.05.014},
  url = {https://www.sciencedirect.com/science/article/pii/S0898122120302182},
  author = {Atgeirr Fl{\o} Rasmussen and Tor Harald Sandve and Kai Bao and Andreas Lauser and Joakim Hove and B{\aa}rd Skaflestad and Robert Kl{\"o}fkorn and Markus Blatt and Alf Birger Rustad and Ove S{\ae}vareid and Knut-Andreas Lie and Andreas Thune}
}

@article{nordbotten202411th,
  title={The 11th Society of Petroleum Engineers Comparative Solution Project: Problem Definition},
  author={Nordbotten, Jan M and Ferno, Martin A and Flemisch, Bernd and Kovscek, Anthony R and Lie, Knut-Andreas},
  journal={SPE Journal},
  volume={29},
  number={05},
  pages={2507--2524},
  year={2024},
  publisher={OnePetro}
}

\end{document}